%% file: main-4page.tex
\documentclass[]{spie}  %

\usepackage{amsmath,amsfonts,amssymb}
\usepackage{graphicx}
\usepackage[colorlinks=true, allcolors=blue]{hyperref}
\usepackage{subcaption}
\usepackage{algorithm}
\usepackage{algpseudocode}
\usepackage{comment}
\usepackage{booktabs}

\title{An Anatomy-guided 3D Segmentation Pipeline for Prostate Surgical Phantoms from CT Scans}

\author[1,4]{Iven Chen}
\author[1,4]{Kuan-I Chung}
\author[1,4]{Ayberk Acar}
\author[2,4]{Susheela Sharma Stern}
\author[3]{Alejandro O. Chara}
\author[1,4]{Jie Ying Wu}
\author[1,4,†]{Daniel Moyer}
\affil[1]{Department of Computer Science, Vanderbilt University, Nashville, TN, USA}
\affil[2]{Department of Mechanical Engineering, Vanderbilt University, Nashville, TN, USA}
\affil[3]{Vanderbilt University Medical Center, Nashville, TN, USA}
\affil[4]{Vanderbilt Institute for Surgery and Engineering, Nashville, TN, USA}
\authorinfo{†Corresponding Author, Daniel Moyer: daniel.moyer@vanderbilt.edu}
\begin{document} 
\maketitle
\input{sections-4page/0-abstract}
\input{sections-4page/1-intro}
\input{sections-4page/2-methods}
\input{sections-4page/3-results-experiments-merged}
\input{sections-4page/4-conclusion}

\newpage
\input{sections-4page/acknowledgement}

\bibliography{report} %
\bibliographystyle{spiebib} %

\newpage
\input{sections-4page/9-appendix}

\end{document}

%% file: sections-4page/0-abstract.tex
\begin{abstract}
Full autonomy has been a long-term goal in robotic surgery and a near-to-medium-term goal for prostatectomy.
Due to the relatively closed structure of the procedure, its standard-of-care robotic/laparoscopic solutions, and procedure economics, prostate procedures are an optimal early target for surgical automation.
Prototyping and testing of possible solutions is done in constructed laboratory phantoms, either synthetic tissues or animal phantoms.
To enable measurement of perceptual and robotic motion error, phantoms undergo volumetric imaging; these images are critical to both a potential deployed automatic system, used in planning motion and treatment (cutting, ablation, suturing, etc.), as well as \emph{post-hoc} assessment of system performance. %
Rapid iteration on these phantom models yields small, inconsistent datasets, while the current workflow for human annotation is often anatomically unreliable and prohibitively time-consuming, rendering deep learning unsuitable at reasonable scales.
To address these limitations, we present an automated 3D segmentation pipeline that uses classical medical image-processing techniques to encode the known anatomy of the phantoms as explicit geometric and intensity priors, which requires no training corpus and is transferred to each new phantom revision without re-annotation.
Across 57 scans spanning multiple phantom revisions, our pipeline recovers the capsule interior at $0.83$ median Dice on 11 contrast-suppressed scans and detects all 32 annotated focal lesions without false positives, completing a full scan in 2--8 minutes on commodity CPU hardware versus 1--2 hours by hand.

\end{abstract}

\keywords{Segmentation, deformable geometry, shape analysis, modality: CT, radiography}

%% file: sections-4page/1-intro.tex
\section{INTRODUCTION}

Prostatectomy is the removal of the prostate in part or in whole. It is performed for two distinct indications, and the phantoms studied here are built for both. In benign prostatic hyperplasia (BPH), the gland enlarges and obstructs the urinary tract, threatening renal function if left untreated, and the tissue is removed in bulk~\cite{goueli2026management,goueli2026procedural}. In localized prostate cancer, a discrete malignant focus is removed, either with the entire gland, or in focal therapy, on its own~\cite{eastham2026localized,slusarczyk2025focal}.
A minimally invasive intervention via robotic or laparoscopic tooling is the standard surgical route for prostatectomy procedures~\cite{goueli2026procedural}, and autonomous robotic surgery is an increasingly promising route towards improving procedure outcomes~\cite{attanasio2021autonomy}. 
Precise spatial perception is crucial for robotic systems to navigate prostate anatomy and guide interventions. At the same time, 3D segmentation is also used in the laboratory development phase of these autonomous systems to evaluate the performance of both perception and automation systems. For both, the task is to partition each scan into four interlocking structures (lumen, capsule shell, lobe tissue, and focal lesions), where an error in one induces cascading errors in its neighbors.

Contemporary datasets are small and inconsistent even for phantom segmentations, due to the heterogeneity of possible phantoms and variability in their fabrication.
Further, tissue contrast varies between scans, where at the low-contrast end components are nearly indistinguishable
(Figure~\ref{fig:overview}). 
Manual annotation is still prohibitively slow, taking 1--2 hours per scan.
Each of these limitations makes data-hungry supervised learning models such as the common U-Net and successor models ~\cite{isensee2021nnunet, wasserthal2023totalsegmentator} unsuitable for this environment.
Therefore, we propose a fully deterministic, rule-based 3D segmentation pipeline that requires no training corpus and utilizes classical medical image-processing techniques, encoding known anatomy as geometric and intensity-based priors to generate multi-label segmentation. 
This work forms part of the development of an autonomous robotic endoscopy system for trans-urethral prostate intervention, in which accurate 3D segmentation is necessary to supply the spatial information required for procedural planning, depth estimation, mesh generation, tissue simulation, and outcome comparison.

\newlength{\ovh}
\setlength{\ovh}{0.334\linewidth}

\begin{figure}[t]
    \centering
    \begin{subfigure}[b]{0.328\linewidth}
        \centering
        \includegraphics[height=\ovh]{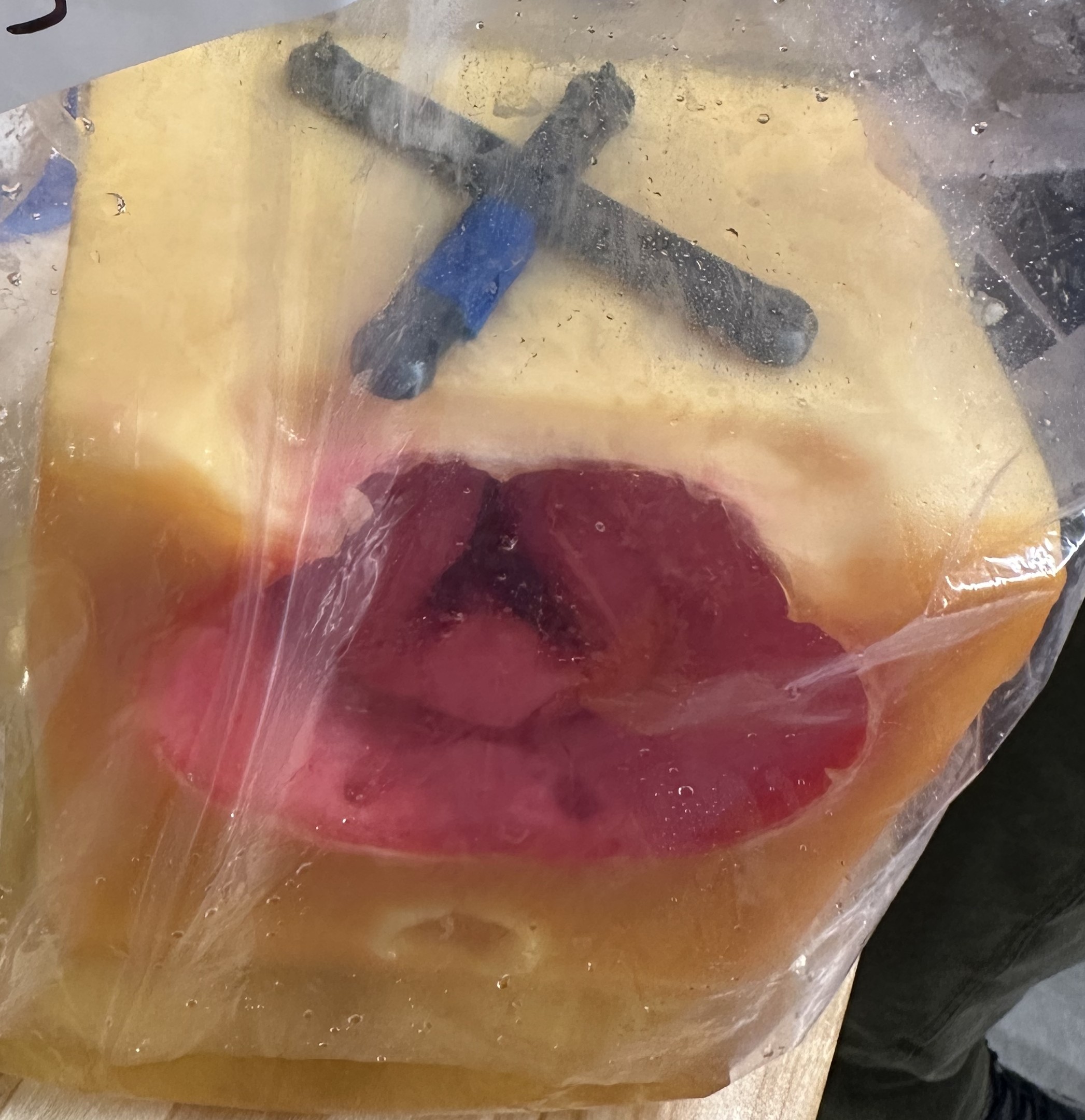}
        \caption{}
        \label{fig:phantom}
    \end{subfigure}\hfill
    \begin{subfigure}[b]{0.652\linewidth}
        \centering
        \includegraphics[height=\ovh]{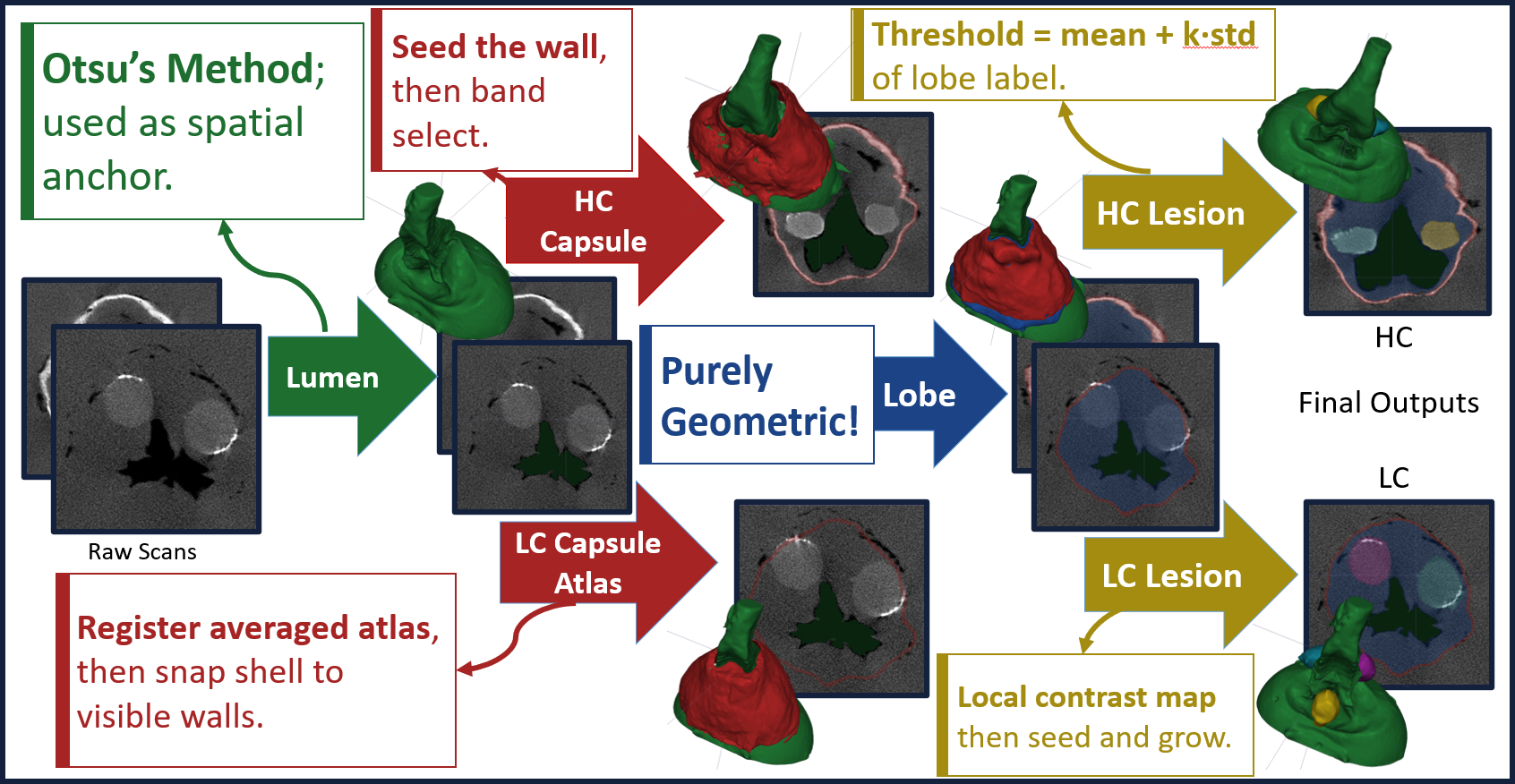}
        \caption{}
        \label{fig:pipeline}
    \end{subfigure}
    \caption{ \textbf{The phantom and the pipeline that segments it.}
    \textbf{(a)}~A physical prostate surgical phantom, with NDI trackers and fiducial markers affixed.
    \textbf{(b)}~Stage chain, with representative volumes at each stage; the high- and low-contrast raw scans at left differ in whether the capsule wall is visible at all.}
    \label{fig:overview}
\end{figure}

%% file: sections-4page/2-methods.tex
\section{METHODS} 

\subsection{Phantom Model and Image Acquisition}

Our prostate phantoms are hydrogel-based tissue models designed for surgical training~\cite{SABA2022100184,790c2f494e294ef0b8ee5deb3f175f25}.
NDI trackers (Polaris; Northern Digital Inc., Waterloo, ON, Canada) are attached to the phantom by tracing the tracker outline onto the phantom surface, excising the corresponding slot with a scalpel, and securing the tracker in place with cyanoacrylate adhesive (Figure~\ref{fig:phantom}). 
Fiducial markers are fixed below the median lobe. 
The phantom is then mounted on a rigid frame, median lobe downward, for consistent positioning, and imaged with a mobile cone-beam CT scanner (xCAT XL; Xoran Technologies, Ann Arbor, MI, USA) at 120 kVp and 57.6 mAs, reconstructed with the vendor's Bone protocol at 0.3 mm isotropic voxel spacing. All series used this single protocol.

\subsection{Pipeline Architecture}

The pipeline is a linear stage chain, in the order from lumen, capsule, lobe, then focal lesions. 
The capsule and lesion stages have high- and low-contrast variants selected by a pipeline-wide contrast flag, while the lumen and lobe stages are contrast-independent and shared (Figure~\ref{fig:pipeline}).
In the final output, labels are mutually exclusive, and conflicts are resolved by a fixed priority order: lumen, lesion, capsule, then lobe, so that structures are cut flush and no voxel carries two labels. The pipeline was implemented in Python using NumPy~\cite{harris2020numpy}, SciPy~\cite{virtanen2020scipy}, scikit-image~\cite{vanderwalt2014scikit}, and SimpleITK~\cite{lowekamp2013simpleitk}.

\subsubsection{Lumen: The Spatial Anchor}

The lumen is obtained by clustering the image histogram using Otsu's Method~\cite{4310076} to extract all air in the scan, which appears as the darkest voxels (around $-$1000 to $-$410 HU). For room air outside of the physical phantom that may have been included, the structure is then geometrically shaved, utilizing a bounding-box circularity and extent as criteria until the cross-section is flat against the phantom model's major external wall (Algorithm~\ref{alg:lumen}). Because the lumen is an air/tissue boundary, it is recovered identically at both contrast levels and thus serves as the spatial anchor for all subsequent stages.

\subsubsection{Capsule: The Prostate Boundary}

\noindent \textbf{High Contrast:} Capsule, lobe, and focal lesions share contrast, so a threshold low enough to catch fainter parts of the wall can flood into the lobe or pick up a lesion. We therefore band-confine~\cite{soille2003morphological} the capsule. Bright components anchored to a ring around the lumen seed the wall, a three-view visual hull~\cite{273735} closes the seed into an interior envelope, and a rind band about that envelope selects the wall itself, rejecting focal lesions and noise for free (Algorithm~\ref{alg:hc-capsule}).

\noindent \textbf{Low Contrast:} As the capsule is essentially invisible here, its location is obtained from an atlas~\cite{iglesias2015multi,klein2008automatic} registered through the lumen (Algorithm~\ref{alg:align-lumen}) and thresholded to an interior envelope. Where thin-wall evidence exists, a ridge detector locates bright wall arcs~\cite{soille2003morphological}, and the boundary is pulled onto them by a regularized displacement that decays back to the atlas where no arc is found, so that unobserved wall is never invented. 
The envelope is then hollowed into a shell and opened at the poles to thread the query lumen through (Algorithm~\ref{alg:lc-capsule}).

\noindent \textbf{Atlas Creation:} The atlas is created from 11 verified, fully segmented high-contrast scans. 
Each capsule is filled into an interior solid, since thin shells average to mush, and all cases are first aligned to a prime lumen via principal component analysis, then averaged~\cite{joshi2004unbiased} using groupwise diffeomorphic demons~\cite{vercauteren2009diffeomorphic} (Algorithm~\ref{alg:make-atlas}).

\subsubsection{Lobe: Defined, not Detected}

The capsule is meant to be the hard geometric boundary, and lobe tissue shares the same HU as tissue outside of the capsule; geometry is used to ensure that lobe voxels do not bleed past the boundary. 
The lobe is defined by filling the capsule interior via a three-view visual hull~\cite{273735}, then subtracting the capsule shell and lumen (Algorithm~\ref{alg:lobe}).

\subsubsection{Focal Lesions}

\noindent \textbf{High Contrast:} With the lobe already labeled, we can obtain an adaptive threshold directly from the distribution within the lobe. 
The lobe core is eroded slightly to avoid contact with the capsule and lumen, then lesions are thresholded by a mean plus a factor times the standard deviation of the lobe's distribution (Algorithm~\ref{alg:hc-lesions}).

\noindent \textbf{Low Contrast:} Here, the capsule is an atlas, so lesions may protrude outside the estimation.
Furthermore, these lesions have an intensity much closer to the lobe. 
Thus, our strategy changes to use the atlas plus some margin to bound our search radius, then apply a local-mean contrast field. 
Where local intensity spikes, these areas are seeded and grown to the full size~\cite{vincent1993morphological}. 
Components that are too small, or whose contrast is inconsistent with lesion tissue, are then discarded (Algorithm~\ref{alg:lc-lesions}). 
Where lesions indeed protrude outside of the capsule wall, the shell is locally stretched outwards via morphology until the lesions are fully contained. 
Lobe fills into the space created by the newly stretched shell, ensuring that lesions are fully enclosed (Algorithm~\ref{alg:lc-reconcile}).

%% file: sections-4page/3-results-experiments-merged.tex
\section{Experiments and Results}

Our corpus consists of 57 CT scans spanning several revisions of the phantom formula, of which 11 are high-contrast, with 3 containing lesions, and the remainder low-contrast.
High-contrast scans are the simpler case and supply reference segmentations rather than being measured, so they are used as a tool to help evaluate the low-contrast path.
The nine annotated low-contrast series used for lesion scoring come from seven distinct phantoms; two of them contribute a second acquisition, taken after the phantom had dried between sessions (Lester) and with a differing number of implanted fiducial markers (Murphy), so the repeats vary imaging conditions rather than duplicating a measurement (Table~\ref{tab:lesions-percase}). 
Each of the two studies below is built around a control that is capable of failing, so that a successful outcome is distinguishable from an uninformative one.

\newsavebox{\resultplot}
\sbox{\resultplot}{\includegraphics[width=0.615\linewidth]%
    {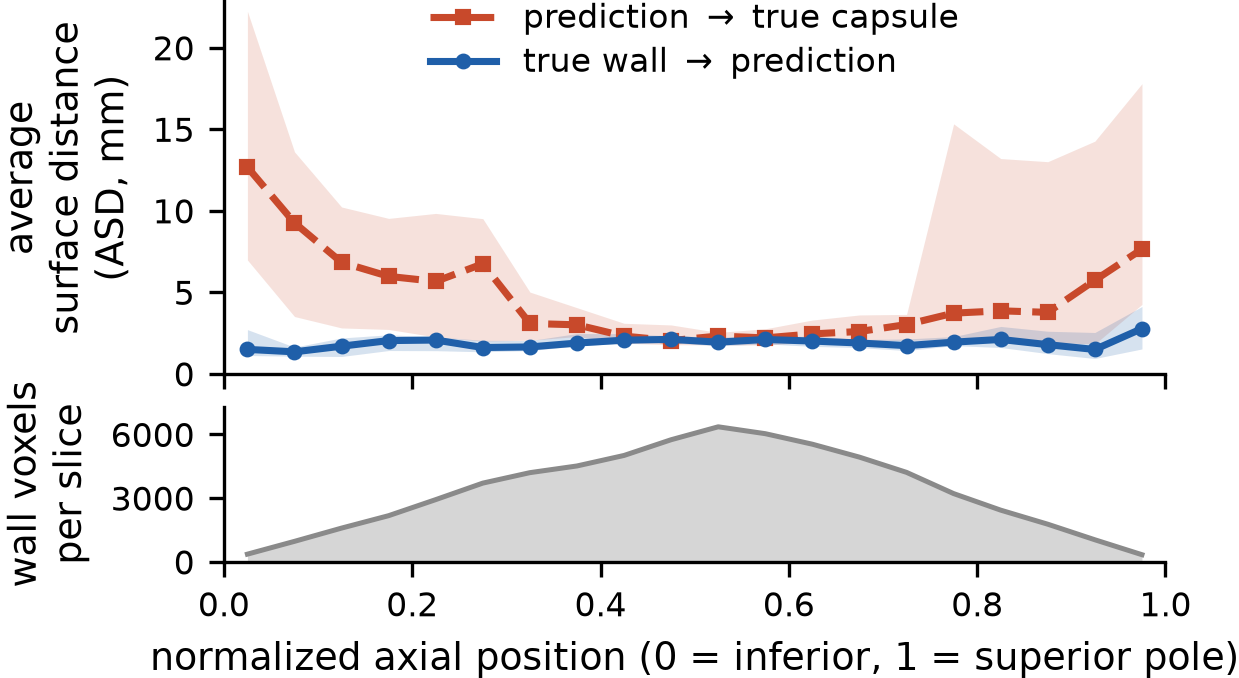}}

\begin{figure}[t]
    \centering
    \begin{subfigure}[b]{0.615\linewidth}
        \centering
        \usebox{\resultplot}
        \caption{Capsule localization after spoofing}
        \label{fig:capsule-profile}
    \end{subfigure}\hfill
    \begin{subfigure}[b]{0.365\linewidth}
        \centering
        \begin{minipage}[c][\ht\resultplot][c]{\linewidth}
        \centering
        \small
        \setlength{\tabcolsep}{4pt}
        \begin{tabular}{lccc}
        \hline
                    & Manual & Spoofed & All \\
        \hline
        Series      & 9      & 3       & 12  \\
        Lesions     & 26     & 6       & 32  \\
        Detected    & 26     & 6       & 32  \\
        False pos.  & 0      & 0       & 0   \\
        \hline
        Median Dice        & 0.911  & 0.932   & --- \\
        Median Recall      & 0.994  & 0.970   & --- \\
        Median Precision   & 0.857  & 0.937   & --- \\
        \hline
        \end{tabular}
        \end{minipage}
        \caption{Focal lesion results}
        \label{tab:lesions}
    \end{subfigure}
    \caption{\textbf{Results of the two studies.} \textbf{(a)}~Capsule localization after contrast spoofing, pooled over the 11 cases and binned into 20 normalized axial positions; ribbons are interquartile ranges. \emph{Top:}  blue is how well the true wall is recovered; red is the predicted surface with no true counterpart. \emph{Bottom:} mean true-wall voxels per slice, indicating where the wall exists to be found. Per-case results are given in Table~\ref{tab:capsule-percase}. \textbf{(b)}~Manual denotes the manually annotated low-contrast series. Dice, recall, and precision are computed once per series over all its lesion voxels; the table reports the median across series. Per-series detail is given in Table~\ref{tab:lesions-percase}.}
    \label{fig:results}
\end{figure}

\subsection{Study 1: Capsule Spoofing}

Low-contrast capsule accuracy cannot be measured directly. 
The wall is absent from the image, so there is no boundary to annotate and no ground truth to score against. 
We therefore synthesized a low-contrast counterpart from each of the 11 high-contrast scans, suppressing the contrast of the capsule voxels by refilling them with the local mean and noise variance of the surrounding soft tissue~\cite{valverde2014filling}, while leaving every other voxel in the volume unchanged, ran the low-contrast path on the synthetic scan, and scored its capsule against the high-contrast segmentation of the original, which serves as ground truth precisely because the wall was visible before it was removed. 
The atlas is built from these same cases, so this measures transfer under a removed wall rather than generalization to an unseen phantom. 
Because the spoof does not alter air, a correct pipeline must reproduce the original lumen exactly; we require bit-identical lumen agreement as the control.

That control passed in every case (lumen Dice $= 1.0000$, 11 of 11), so all capsule disagreement is attributable to the atlas rather than to the spoof. 
Interior Dice~\cite{dice1945measures} was median $0.829$ (range $0.790$--$0.881$), with recall $0.819$ and precision $0.883$. 
Boundary error, reported as a directed pair of per-case average surface distances (ASD)~\cite{taha2015metrics}, was $2.36$ mm from the true capsule to the prediction and $7.91$ mm in the reverse direction; the asymmetry holds in every case (median 3.4×, range 1.8--5.0×) and is over-extension rather than a failure to find the wall (Figure~\ref{fig:results}a). 

\subsection{Study 2: Lesion Recall}

Lesions are scored per matched component~\cite{maierhein2024metrics,reinke2024understanding}: each annotated lesion, obtained manually by user-guided active-contour segmentation in ITK-SNAP~\cite{yushkevich2006itksnap} or by high-contrast spoofing, is paired with the detection of greatest overlap, and recall and Dice are computed on that pair. 
Any component picked up by the detector that does not have an annotated counterpart counts as a false positive. 
Fiducial markers and NDI trackers serve as our controls, as they share lesion characteristics by being bright and compact.

All $32$ annotated lesions were detected, with no false positives on any series (Figure~\ref{fig:results}b).
On the nine manually annotated low-contrast series ($26$ lesions), per-lesion recall was median $0.997$ (mean $0.979$, minimum $0.872$), with $24$ of $26$ lesions at or above $0.90$ recall, and Dice was median $0.976$ (mean $0.937$, minimum $0.730$).
On the three spoofed series ($6$ lesions), scored against the high-contrast lesion masks of the same scans, per-lesion recall was median $0.985$ and Dice median $0.941$.
No fiducial markers or NDI trackers were labeled as a lesion. 

%% file: sections-4page/4-conclusion.tex
\section{CONCLUSION}

We presented a deterministic, training-free pipeline that partitions a prostate surgical phantom scan into four interlocking structures across contrast levels utilizing geometric and intensity-based priors to reconstruct the anatomy. 
What transfers is the organizing principle: segmenting each structure through the observable channel, and supplying a prior only where the image carries no evidence. 
The limitation lies at the same seam. 
Where the wall is invisible, the atlas is the only source of boundary, and it overextends at the poles, where no wall constrains it. 
At 2--8 minutes per scan on commodity CPU hardware, compared with 1--2 hours by hand, the pipeline removes the annotation bottleneck that keeps these datasets too small to learn from.

%% file: sections-4page/acknowledgement.tex
\acknowledgments %
 
Research reported in this publication was supported by the Advanced
Research Projects Agency for Health (ARPA-H) under the ALISS project,
Award Number D24AC00415-00. The ARPA-H award of up to \$11,935,038
provided 100\% of the financial support for this work. The opinions and
findings in this paper are solely the responsibility of the authors and do not necessarily represent the official views of ARPA-H.  

%% file: sections-4page/9-appendix.tex
\appendix

\renewcommand{\thesection}{\Alph{section}}

\renewcommand{\thefigure}{\thesection\arabic{figure}}
\setcounter{figure}{0}

\renewcommand{\thetable}{\thesection\arabic{table}}
\setcounter{table}{0}

\renewcommand{\theequation}{\thesection\arabic{equation}}
\setcounter{equation}{0}

\renewcommand{\thealgorithm}{\thesection\arabic{algorithm}}
\setcounter{algorithm}{0}

\section{Algorithms}
\input{assets/algorithms/algo}

\clearpage
\section{Data}
\setcounter{table}{0}
\input{assets/data/data}

%% file: assets/algorithms/algo.tex
\begin{algorithm}
\caption{Lumen mask --- the contrast-independent anchor}
\label{alg:lumen}
\begin{algorithmic}[1]
\Require CT volume $V$
\Ensure  Lumen mask $L$
\State $T \gets V > \textsc{Otsu}(V)$
      \Comment{air/tissue split, contrast-independent}
\State $L \gets \textsc{LargestCC}\big(\neg T \cap \textsc{FillHoles}(T)\big)$
      \Comment{air enclosed by tissue}
\State shave end slices off each face of $L$ until the cross-section meets its
       shape target \Comment{removes the slack air gaps}
\State \textbf{return} $\textsc{Close}(\textsc{LargestCC}(\textsc{Open}(L)))$
\end{algorithmic}
\end{algorithm}
\noindent\emph{Why it anchors everything:} the shave removes whole axial slices, so
the two ends of $L$ are $z$-perpendicular flat faces that mark the phantom's true
axial extent --- reused by every later stage.

\begin{algorithm}
\caption{Capsule mask, high contrast --- envelope--band selection}
\label{alg:hc-capsule}
\begin{algorithmic}[1]
\Require $V$; lumen $L$; bright level $h_s$; faint level $h_t$;
         trim level $h_{\mathrm{tr}}$ and edge level $\gamma$
\Ensure  Capsule mask $C$; interior envelope $E$
\State $S \gets$ components of $V \ge h_s$ that touch $\textsc{Dilate}(L) \setminus L$
      \Comment{seed the wall off the lumen ring}
\State $E \gets \textsc{VisualHull}(S)$
\State $\mathcal{B} \gets \textsc{Dilate}(E) \setminus \textsc{Erode}(E)$
      \Comment{rind band about the envelope surface}
\State $C \gets$ every component of $V \ge h_s$ that touches $\mathcal{B}$
      \Comment{\emph{whole} components, not $(V \ge h_s) \cap \mathcal{B}$}
\State $C \gets C \;\cup\; \big[(h_t \le V < h_s) \cap \mathcal{B} \cap \textsc{Dilate}(C)\big]$
      \Comment{faint wall, distance-constrained}
\State $C \gets C \;\cup\; \big(\mathcal{B} \cap \textsc{Close}(C)\big)$
      \Comment{bridge gaps; the poles stay open}
\State $C' \gets C \cap \big[(V \ge h_{\mathrm{tr}}) \,\cup\, (\lVert \nabla G_\sigma \! * \! V \rVert \ge \gamma)\big]$
      \Comment{edge trim: keep bright \textbf{or} wall-edge}
\State \textbf{return} $\textsc{Close}(C') \cap C$, \; $E$
\end{algorithmic}
\end{algorithm}
\noindent\emph{Three load-bearing choices.} Capsule and lobe are HU-connected (the lobe
reaches ${\sim}464$\,HU), so every recovery step must be band-confined --- a global cut at
$h_t$ floods straight through the lobe. Selecting \emph{whole} components rather than
intersecting with $\mathcal{B}$ matters: the intersection discards ${\sim}61\%$ of the true
wall, whereas whole components keep each arc at its own thickness and reject interior
lesions for free. And the trim's \emph{edge} term preserves the faint true wall a pure-HU
cut would eat; the re-close is intersected back with the pre-trim mask, so it can never
bulge into lobe or exterior.

\begin{algorithm}
\caption{Lobe mask --- defined geometrically, not detected}
\label{alg:lobe}
\begin{algorithmic}[1]
\Require Capsule $C$; lumen $L$; raw $V$; envelope $E$; pole tissue floor $h_p$
\Ensure  Lobe mask $G$
\State $N \gets \neg(C \cup L)$
      \Comment{capsule and lumen outrank the lobe}
\State $G \gets E \cap N$
      \Comment{$\mathrm{lobe} = \mathrm{fill(interior)} - \mathrm{capsule} - \mathrm{lumen}$}
\State $G \gets \textsc{LargestCC}(\textsc{Open}(G))$
      \Comment{severs the thin necks leaks escape through}
\State $G \gets \textsc{Close}(G) \cap N \cap E$
      \Comment{smooth, then re-cut}
\State $G \gets G \;\cup\; \big(\textsc{FillHoles}(G \cup L) \cap N \cap \neg E \cap (V > h_p)\big)$
      \Comment{pole protrusion, outside the envelope}
\State \textbf{return} $\textsc{ClipToDepthSpan}(G, L)$
\end{algorithmic}
\end{algorithm}
\noindent\emph{Why geometry and not intensity:} the lobe cannot be separated from the
capsule by HU, but it can be defined by containment --- everything inside the capsule
that is not lumen \emph{is} lobe, so imaging artifacts inside the boundary are included
automatically. Two steps guard that definition: the re-cut, because the closing can bulge
past the wall; and the pole step, because real tissue exits the open poles \emph{outside}
the envelope where every interior-clipped step misses it, with the HU floor admitting that
tissue while rejecting the slack air gap beside it.

\newpage

\begin{algorithm}
\caption{Focal lesions, high contrast}
\label{alg:hc-lesions}
\begin{algorithmic}[1]
\Require $V$; lobe $G$; factor $k$; size floor $n_{\min}$; roundness floor $\psi$; count cap $\kappa$
\Ensure  Lesion label map $X$
\State $G_c \gets \textsc{Erode}(G)$
      \Comment{core: drops the bright inner-wall rim}
\State $\tau \gets \mathrm{mean}_{G_c}(V) + k \cdot \mathrm{std}_{G_c}(V)$
      \Comment{adaptive: the lobe's own bright tail}
\State $M \gets (V \ge \tau) \cap G_c$
\State $M \gets \textsc{Reconstruct}\big(\textsc{Erode}(M),\, M\big)$
      \Comment{opening \emph{by reconstruction}}
\State $\mathcal{C} \gets \{\omega \in \mathrm{CC}(\textsc{FillHoles}(\textsc{Close}(M))) :
        |\omega| \ge n_{\min} \ \wedge\ \textsc{Roundness}(\omega) \ge \psi\}$
      \Comment{gate the \textbf{raw} component}
\State \textbf{return} the $\kappa$ largest of $\mathcal{C}$, labeled largest-first
       without overwrite, each passed through \textsc{Regularize}
       (Algorithm~\ref{alg:regularize})
\end{algorithmic}
\end{algorithm}
\noindent\emph{Three choices that matter.} The core erosion excludes the bright
partial-volume rim on the inner capsule wall, which otherwise both inflates $\tau$
\emph{and} thresholds in as a false lesion. Reconstruction replaces a plain opening,
which erased a real $31$k-voxel lesion outright. And the gate reads the \emph{raw}
component because \textsc{Regularize} raises roundness --- gating after it would let an
elongated artifact round up into a false lesion.

\begin{algorithm}
\caption{\textsc{Regularize} --- per-lesion shape regularization}
\label{alg:regularize}
\begin{algorithmic}[1]
\Require Component $\omega$; ellipsoid factor $k$; hull votes $v$; median radius $m$
\Ensure  Regularized component $\omega'$
\State fit the centroid and covariance of $\omega$, refitting three times on the
       inliers $\{x : d^2_M(x) \le 5\}$
      \Comment{robust: an artifact tail cannot stretch the axes}
\State $\omega' \gets \omega \cap \{x : d^2_M(x) \le 5k^2\}$
      \Comment{ellipsoid cap --- a pure trim}
\State $\omega' \gets \{x :$ the per-slice convex hull of $\omega'$ covers $x$ in
       $\ge v$ of the three axis views$\}$
      \Comment{fills craters \textsc{FillHoles} cannot reach}
\State \textbf{return} $\textsc{Median}(\omega', m)$
      \Comment{round the surface, trim thin spikes}
\end{algorithmic}
\end{algorithm}
\noindent\emph{Why cap before hull, and why one setting fits every lesion:} the cap runs
first so the convex hull cannot bridge across an elongated CT streak and re-absorb it.
Both steps are self-limiting --- a genuinely round lesion lies inside its own ellipsoid
and equals its own hull, so neither can inflate it --- which is why the same three values
serve both contrast levels. The hull is what recovers a heterogeneous lesion's hollow
core, an \emph{open} crater that no hole-filling can reach. The cap is a no-op below
${\sim}200$ voxels, where the covariance is unstable.

\begin{algorithm}
\caption{\textsc{AlignLumen} --- lumen-referenced similarity alignment}
\label{alg:align-lumen}
\begin{algorithmic}[1]
\Require Moving lumen $L_m$; fixed lumen $L_f$
\Ensure  Similarity transform $T^\star$
\State $\mathcal{T} \gets \{\text{the four proper axis flips of the PCA frame } L_m \! \to \! L_f\}$
\State $\mathcal{T} \gets \mathcal{T} \cup \{\text{the identity pose: uniform scale and centroid only}\}$
      \Comment{covers the roll-degenerate lumen}
\For{$T \in \mathcal{T}$}
    \State $T' \gets$ similarity registration seeded at $T$
           \Comment{seeding avoids the wrong-flip basin}
    \State keep $T'$ only if $\textsc{Dice}(T'(L_m), L_f) \ge \textsc{Dice}(T(L_m), L_f)$
           \Comment{guard: cannot worsen a GT-less scan}
\EndFor
\State \textbf{return} the candidate of highest \textbf{final} Dice
\end{algorithmic}
\end{algorithm}
\noindent\emph{Why the final score and not the flip score:} the flips enumerate only discrete
$180^\circ$ poses, so a residual continuous roll can survive the choice, and on a
roll-degenerate lumen --- two largest eigenvalues near-equal --- the PCA axes and therefore
all four flips are ill-determined. Ranking \emph{after} refinement is what makes the
selection regression-free; the spread of the flip scores is retained as a free confidence
signal at apply time.

\begin{algorithm}
\caption{Capsule atlas construction (offline, high-contrast corpus)}
\label{alg:make-atlas}
\begin{algorithmic}[1]
\Require High-contrast cases $\{V_i\}$; prime case index $p$
\Ensure  Occupancy map $\Pi$ over the prime grid
\For{each case $i$}
    \State $L_i \gets$ Algorithm~\ref{alg:lumen}$(V_i)$;\quad
           $(C_i, E_i) \gets$ Algorithm~\ref{alg:hc-capsule}$(V_i, L_i)$
    \Comment{store the filled interior, never the shell}
\EndFor
\State $\bar{E} \gets \dfrac{1}{n}\displaystyle\sum_i \textsc{AlignLumen}(L_i, L_p)(E_i)$
       \Comment{align through the lumen}
\Repeat
    \State warp every aligned $E_i$ onto $\bar{E}$ by diffeomorphic demons on interior
           signed-distance maps, and re-average
\Until{converged}
\State \textbf{return} $\Pi \gets \bar{E}$
\end{algorithmic}
\end{algorithm}
\noindent\emph{Why deformable here but not at apply time:} shells average to mush, so the
filled interior is stored; and because every case is a segmented interior rather than an
image, the warp is legitimate and sharpens the consensus (fuzzy band
$44\% \! \to \! 13\%$). At apply time only the lumen is observable, so
Algorithm~\ref{alg:lc-capsule} stays similarity-only.

\begin{algorithm}
\caption{Capsule mask, low contrast --- atlas transfer with arc snap}
\label{alg:lc-capsule}
\begin{algorithmic}[1]
\Require $V$; lumen $L$; atlas $\Pi$ over prime lumen $L_p$; occupancy cut $\theta$;
         shell thickness $t$; pole clearance $c$
\Ensure  Capsule mask $C$; interior envelope $E$
\State $E \gets \textsc{Resample}\big(\Pi,\ \textsc{AlignLumen}(L_p, L)\big) \ge \theta$
      \Comment{transferred interior envelope}
\State $A \gets$ coherent thin bright ridges near $\partial E$
      \Comment{white top-hat, locally $z$-scored}
\State $E \gets \textsc{SnapBoundary}(E, A)$
      \Comment{regularized pull of $\partial E$ onto the arcs}
\State $C \gets \big(E \setminus \textsc{Erode}(E,t)\big) \setminus \textsc{Dilate}(L,c)$
      \Comment{opens both poles; enforces lumen $>$ capsule}
\State \textbf{return} $\textsc{LargestCC}(C)$, \; $E$
\end{algorithmic}
\end{algorithm}
\noindent\emph{Evidence-gated by construction.} A \emph{ridge} detector is used so broad
bright lobe and partial-volume regions are ignored while faint thin walls survive. The snap
decays back to the atlas away from the arcs, so where no wall is seen the boundary stays
where the atlas put it and unobserved wall is never invented.

\begin{algorithm}
\caption{Focal lesions, low contrast --- tissue-only local contrast}
\label{alg:lc-lesions}
\begin{algorithmic}[1]
\Require $V$; envelope $E$; lumen $L$; box half-width $R$; tissue floor $h_f$;
         seed level $\sigma_s$ and grow level $\sigma_g < \sigma_s$;
         contrast band $[\beta_{\mathrm{lo}}, \beta_{\mathrm{hi}}]$; size floor $n_{\min}$; cap $\kappa$
\Ensure  Lesion label map $X$
\State $\Theta \gets V > h_f$
      \Comment{tissue}
\State $\displaystyle \Phi \gets V - \frac{\textsc{BoxMean}(V \cdot \Theta,\, R)}{\textsc{BoxMean}(\Theta,\, R)}$
      \Comment{local contrast against tissue alone}
\State $\mathcal{O} \gets \{\,$within $\delta$ of $E\,\} \setminus \textsc{Dilate}(L)$
      \Comment{search region}
\State $\mathcal{S} \gets$ components of $(\Phi > \sigma_s) \cap \mathcal{O}$ above the seed floor
\State $M \gets$ components of $(\Phi > \sigma_g) \cap \mathcal{O}$ that meet $\mathcal{S}$
      \Comment{grow, to fill the lesion interior}
\State $\mathcal{C} \gets \{\omega \in \mathrm{CC}(M) : |\omega| \ge n_{\min} \ \wedge\
        \beta_{\mathrm{lo}} \le \mathrm{median}\,\Phi[\omega] \le \beta_{\mathrm{hi}}\}$
      \Comment{shape-neutral gate, deliberately not roundness}
\State \textbf{return} the $\kappa$ largest of $\mathcal{C}$, labeled as in
       Algorithm~\ref{alg:hc-lesions}
\end{algorithmic}
\end{algorithm}
\noindent\emph{Two constraints a reimplementation must keep.} $R$ must \emph{exceed} the
largest lesion: with a smaller box the local mean tracks the lesion itself and its interior
contrast vanishes, and excluding air keeps the lumen from dragging the background estimate
down. The gate is shape-neutral because these lesions are genuinely oblong --- the floor
drops marginal lobe blobs, the ceiling drops ultra-bright compact vessels and
calcifications.

\begin{algorithm}
\caption{Lesion reconcile, low contrast --- grow the geometry to the lesion}
\label{alg:lc-reconcile}
\begin{algorithmic}[1]
\Require Pristine capsule $C$, interior $E$, lobe $G$; lumen $L$; lesion map $X$;
         lobe pad $p$; shell thickness $t$; pole clearance $c$; stretch spread $s$
\Ensure  Reconciled capsule $C'$, interior $E'$, lobe $G'$
\State \textbf{if} every $\omega \subseteq X$ lies in $G$ with a $\ge p$ collar from $C$
       \textbf{then return} $C,\, E,\, G$ unchanged
\State $\Delta \gets \max_{\omega \subseteq X}\ \textsc{Smoothstep}(1 - d_\omega / s)\cdot D_\omega$,
       \quad $D_\omega = \max_{\omega}\textsc{SignedDist}(E) + (p + t)$
      \Comment{full at the lesion, zero at $s$}
\State $E' \gets \textsc{FillHoles}\big(\,(\textsc{SignedDist}(E) \le \Delta)
       \cup \textsc{Dilate}(X,\, p + t) \cup L\,\big)$
      \Comment{stretch, contain, close pockets}
\State $\mathcal{R} \gets \textsc{Dilate}\big((E' \setminus E) \cup (X \setminus \textsc{Erode}(E, p))\big)$
      \Comment{neighborhood of change}
\State $C' \gets (C \setminus \mathcal{R}) \;\cup\; \big((E' \setminus \textsc{Erode}(E',t)) \cap \mathcal{R}\big)$
      \Comment{re-shell locally, splice into the global wall}
\State $C' \gets C' \setminus \textsc{Dilate}(L,\, c) \setminus \textsc{Dilate}(X,\, p-1)$
      \Comment{reopen the poles; hold the lobe collar}
\State $G' \gets \big(G \cup X \cup (E' \cap \mathcal{R})\big) \setminus C' \setminus L$
      \Comment{absorb the lesion and its collar}
\State \textbf{return} $\textsc{LargestCC}(C')$, \; $E'$, \; $\textsc{LargestCC}(G')$
\end{algorithmic}
\end{algorithm}
\noindent\emph{Why the geometry moves and not the lesion:} the capsule here is a registered
prior while the lesion is read from the image, so where the two disagree it is the boundary
that should move --- clipping the lesion would discard observed tissue to preserve an
estimate. The falloff keeps that push local: a uniform one inflates the whole gland, a hard
one leaves a bump, and a lesion already deep inside gives $D_\omega \le 0$ and contributes
nothing. Reading the pristine masks and writing separate reconciled ones makes the stage
idempotent, so the stretch can never compound.

\begin{table}[t]
\centering
\caption{Parameter values. Radii and half-widths are in voxels unless marked mm.}
\label{tab:bph-params}
\small
\begin{tabular}{@{}llr@{}}
\toprule
Alg. & Parameter & Value \\
\midrule
\ref{alg:lumen} & morphology radii (open, close)   & $2,\,2$ \\
\midrule
\ref{alg:hc-capsule} & bright level $h_s$          & $300$ HU \\
   & faint level $h_t$ / growth                & $240$ HU / $3$ \\
   & band depth (inner, outer)                 & $6,\,2$ \\
   & trim level $h_{\mathrm{tr}}$ / edge $\gamma$ / $\sigma$ & $240$ HU / $120$ / $1.5$ \\
   & closing radii (band, planar, trim)        & $5,\,30,\,10$ \\
\midrule
\ref{alg:lobe} & opening / closing radius        & $5,\,5$ \\
   & pole tissue floor $h_p$                   & $-300$ HU \\
\midrule
\ref{alg:hc-lesions} & core erosion / factor $k$ & $1$ / $1.0$ \\
   & size floor $n_{\min}$ / roundness $\psi$  & $8000$ / $0.6$ \\
\midrule
\ref{alg:regularize} & ellipsoid $k$, hull votes, median & $1.25,\ 3,\ 2$ \\
\midrule
\ref{alg:make-atlas} & corpus size $n$ / groupwise sweeps & $11$ / $3$ \\
   & deformable working resolution             & $2.0$ mm iso \\
\midrule
\ref{alg:lc-capsule} & occupancy cut $\theta$ / shell $t$ & $0.5$ / $3$ \\
   & pole clearance $c$                        & $2$ \\
   & top-hat scale / $z$ / min.\ arc size      & $9$ / $1.2$ / $100$ \\
   & snap reach / clamp / spread               & $12$ / $20$ / $2.5$ (mm) \\
\midrule
\ref{alg:lc-lesions} & box half-width $R$ / tissue floor $h_f$ & $40$ / $-200$ HU \\
   & seed $\sigma_s$ / grow $\sigma_g$         & $110$ / $60$ \\
   & contrast band $[\beta_{\mathrm{lo}},\beta_{\mathrm{hi}}]$ & $[85,\,170]$ \\
   & size floor $n_{\min}$ / ROI margin $\delta$ & $3000$ / $25$ \\
\midrule
\ref{alg:hc-lesions},\,\ref{alg:lc-lesions} & lesion count cap $\kappa$ & $3$ \\
\midrule
\ref{alg:lc-reconcile} & lobe pad $p$ / stretch spread $s$ & $3$ / $20$ mm \\
   & de-burr opening / closing radius           & $1,\,1$ \\
\bottomrule
\end{tabular}
\end{table}

%% file: assets/data/data.tex
\begin{table}[ht]
\centering
\caption{Per-series focal lesion results. Dice, recall, and precision are computed on the union of lesions within each series. The suppressed series are high-contrast scans passed through the low-contrast path, scored against the high-contrast segmentation of the same scan. Lester and Murphy each appear twice, numbered in acquisition order, as two separate scanning sessions of the same phantom.}
\label{tab:lesions-percase}
\begin{tabular}{lccccc}
\hline
Series & GT lesions & Detected & Dice & Recall & Precision \\
\hline
\multicolumn{6}{l}{\emph{Low contrast, manual annotation}} \\
Leo               & 3 & 3 & 0.894 & 0.940 & 0.853 \\
Larry             & 3 & 3 & 0.860 & 0.948 & 0.786 \\
Lester 1          & 3 & 3 & 0.890 & 0.994 & 0.806 \\
Lester 2          & 3 & 3 & 0.911 & 0.973 & 0.857 \\
Louis             & 2 & 2 & 0.896 & 0.956 & 0.843 \\
Matthew           & 3 & 3 & 0.990 & 0.995 & 0.984 \\
Monty             & 3 & 3 & 0.988 & 0.996 & 0.981 \\
Murphy 1          & 3 & 3 & 0.990 & 1.000 & 0.981 \\
Murphy 2          & 3 & 3 & 0.987 & 1.000 & 0.975 \\
Median            & --- & 26/26 & 0.911 & 0.994 & 0.857 \\
\hline
\multicolumn{6}{l}{\emph{Suppressed low-contrast}} \\
Suppressed 1         & 2 & 2 & 0.911 & 0.998 & 0.838 \\
Suppressed 2         & 2 & 2 & 0.932 & 0.927 & 0.937 \\
Suppressed 3         & 2 & 2 & 0.982 & 0.970 & 0.994 \\
Median            & --- & 6/6 & 0.932 & 0.970 & 0.937 \\
\hline
\end{tabular}
\end{table}

\begin{table}[ht]
\centering
\caption{Per-case results for the capsule suppression study, ordered by descending interior Dice.
Volumes are of the capsule interior. Average surface distance (ASD) is the mean, over every voxel of one boundary, of the Euclidean distance to the nearest voxel of the other; it is reported as a directed pair because the two directions measure different things: GT→pred. measures how well the true capsule is covered; pred.→GT measures predicted surface having no true counterpart. HD95 is the 95th-percentile Hausdorff distance over the same distances, reported in preference to the maximum, which a single voxel dominates. Lumen Dice is the validity control, not a
result: the suppression does not alter air, so a correct pipeline must reproduce the original lumen exactly.}
\label{tab:capsule-percase}
\small
\begin{tabular}{lrrccccccc}
\hline
 & GT & Pred. & & & & GT $\rightarrow$ pred. & Pred. $\rightarrow$ GT & HD95 & Lumen \\
Case & (mL) & (mL) & Dice & Recall & Prec. & (mm) & (mm) & (mm) & Dice \\
\hline
Case 1     & 182.5 & 181.6 & 0.881 & 0.878 & 0.883 & 1.55 & 7.15 & 28.1 & 1.0000 \\
Case 2     & 181.7 & 159.2 & 0.873 & 0.819 & 0.934 & 1.74 & 5.96 & 22.4 & 1.0000 \\
Case 3     & 151.5 & 157.3 & 0.847 & 0.863 & 0.831 & 1.85 & 9.04 & 35.9 & 1.0000 \\
Case 4     & 167.0 & 169.8 & 0.847 & 0.854 & 0.839 & 1.82 & 9.01 & 32.8 & 1.0000 \\
Case 5     & 148.3 & 159.5 & 0.841 & 0.873 & 0.811 & 1.95 & 8.87 & 34.6 & 1.0000 \\
Case 6     & 189.9 & 144.7 & 0.829 & 0.730 & 0.958 & 2.54 & 4.47 & 10.5 & 1.0000 \\
Case 7     & 207.1 & 152.9 & 0.823 & 0.715 & 0.969 & 3.09 & 5.71 & 19.5 & 1.0000 \\
Case 8     & 174.2 & 164.7 & 0.813 & 0.790 & 0.836 & 2.36 & 7.91 & 26.7 & 1.0000 \\
Case 9     & 205.4 & 160.0 & 0.804 & 0.715 & 0.919 & 2.66 & 5.06 & 13.3 & 1.0000 \\
Case 10    & 208.0 & 156.6 & 0.800 & 0.701 & 0.932 & 2.89 & 8.13 & 26.8 & 1.0000 \\
Case 11    & 139.3 & 168.5 & 0.790 & 0.872 & 0.721 & 2.56 & 12.02 & 36.9 & 1.0000 \\
\hline
Median     & 181.7 & 159.5 & 0.829 & 0.819 & 0.883 & 2.36 & 7.91 & 26.8 & 1.0000 \\
\hline
\end{tabular}
\end{table}